\documentclass[runningheads]{llncs}
\usepackage{makeidx}  
\usepackage{graphicx}
\usepackage{cite}
\usepackage{cleveref}
\usepackage{amsmath}
\usepackage[bottom]{footmisc}

\DeclareMathOperator*{\argmin}{\arg\!\min}
\newcommand{\norm}[1]{\left\lVert#1\right\rVert}
\def\t{\mathbf{t}}
\def\n{\mathbf{n}}
\def\p{\mathbf{p}}

\titlerunning{Real-Time Segmentation of Surgical Tools based on Deep Learning}  
\title{Real-Time Segmentation of Non-Rigid Surgical Tools based on Deep Learning and Tracking}
\author{Luis C. Garc\'ia-Peraza-Herrera\inst{1} \and Wenqi Li\inst{1} \and Caspar Gruijthuijsen\inst{4} \and Alain Devreker\inst{4} \and George Attilakos\inst{3} \and Jan Deprest\inst{5} \and Emmanuel Vander Poorten\inst{4} \and Danail Stoyanov\inst{2} \and Tom Vercauteren\inst{1} \and S\'ebastien Ourselin\inst{1}}
\authorrunning{Luis C. Garc\'ia-Peraza-Herrera et al.} 
\institute{
Translational Imaging Group, CMIC, University College London, UK\\
\and
Surgical Robot Vision Group, CMIC, University College London, UK\\
\and
University College London Hospitals, UK\\
\and
Katholieke Universiteit Leuven, Belgium\\
\and
Universitair Ziekenhuis Leuven, Belgium\\
}
\begin{document}
\maketitle              
\begin{abstract}
Real-time tool segmentation is an essential component in computer-assisted surgical systems. We propose a novel real-time automatic method based on Fully Convolutional Networks (FCN) and optical flow tracking. Our method exploits the ability of deep neural networks to produce accurate segmentations of highly deformable parts along with the high speed of optical flow. Furthermore, the pre-trained FCN can be fine-tuned on a small amount of medical images without the need to hand-craft features. We validated our method using existing and new benchmark datasets, covering both \textit{ex vivo} and \textit{in vivo} real clinical cases where different surgical instruments are employed. Two versions of the method are presented, non-real-time and real-time. The former, using only deep learning, achieves a balanced accuracy of 89.6\% on a real clinical dataset, outperforming the (non-real-time) state of the art by 3.8 percentage points. The latter, a combination of deep learning with optical flow tracking, yields an average balanced accuracy of 78.2\% across all the validated datasets.

\end{abstract}

\section{Introduction}
Tool detection, segmentation and tracking is a core technology that has many potential applications. It may for example be used to increase the context-awareness of surgeons in the operating room~\cite{Bouget2015}.
In the context of delicate surgical interventions, such as fetal~\cite{Daga2015} and ophthalmic surgery~\cite{Sznitman2012}, providing the clinical operator with accurate real-time information about the surgical tools could be highly valuable and help to avoid human errors.
Identifying tools is also part of other computational pipelines such as mosaicking, visual servoing and skills assessment. 
Image mosaicking can provide reconstructions larger than the image provided by the usual endoscopic view. The mosaic is normally generated by stitching endoscopic images as the endoscope moves across the operating site~\cite{Tella2016}. However, surgical tools present in the images occlude the surgical scene being reconstructed.
Real-time instrument detection and tracking facilitates the localisation of the instruments and the further separation from the underlying tissue, so that the final mosaic only contains patient's tissue.
Another application of tool segmentation is visual servoing of articulated or flexible surgical robots. As the dexterity of the instruments rises~\cite{Devreker2015}, it becomes increasingly difficult for the surgeon to understand the shape of these instruments.
With the miniaturisation of said instruments, the kinematics of these devices become less deterministic due to effects from friction, hysteresis and backlash alongside with increased instrument compliance and safety. Furthermore, it is challenging to embed position or shape sensing on them without increasing their size. 
A key advantage of visual tool tracking versus fiducial markers or auxiliary technologies is that there is no need to modify the current workflow or propose alternative exotic instruments. 
Previous work has addressed detection~\cite{Reiter2012a}, localisation~\cite{Allan2013} and pose estimation of instruments~\cite{Allan2014} using different cues and classification strategies. For example, employing information about the geometry of the instruments~\cite{Pezzementi2009}, fiducial markers~\cite{Reiter2011}, 3D coordinates of the insertion point~\cite{Voros2006}, fusing visual and kinematic information~\cite{Reiter2014} and through multi-class pixel-wise classification of colour, texture and position features with different machine learning techniques such as Random Forests (RF)~\cite{Allan2013a} and Boosted Decision Forests~\cite{Bouget2015}. Recent advances in Region-based Convolutional Neural Networks (R-CNN)~\cite{Girshick2015} and Region Proposal Networks (RPN)~\cite{Ren2015} have enabled the possibility of object detection (with a bounding box) near real-time (17fps for images on Pascal VOC 2007~\cite{pascal-voc-2007}).  EndoNet~\cite{endonet} has been recently proposed as a solution for phase recognition and tool presence detection on laparoscopic videos. However, there is still a need for an automatically initialised real-time (i.e. camera frame rate) segmentation algorithm for non-rigid tools with unknown geometry and kinematics.

\begin{figure}[b!]
	\includegraphics[width=\textwidth]{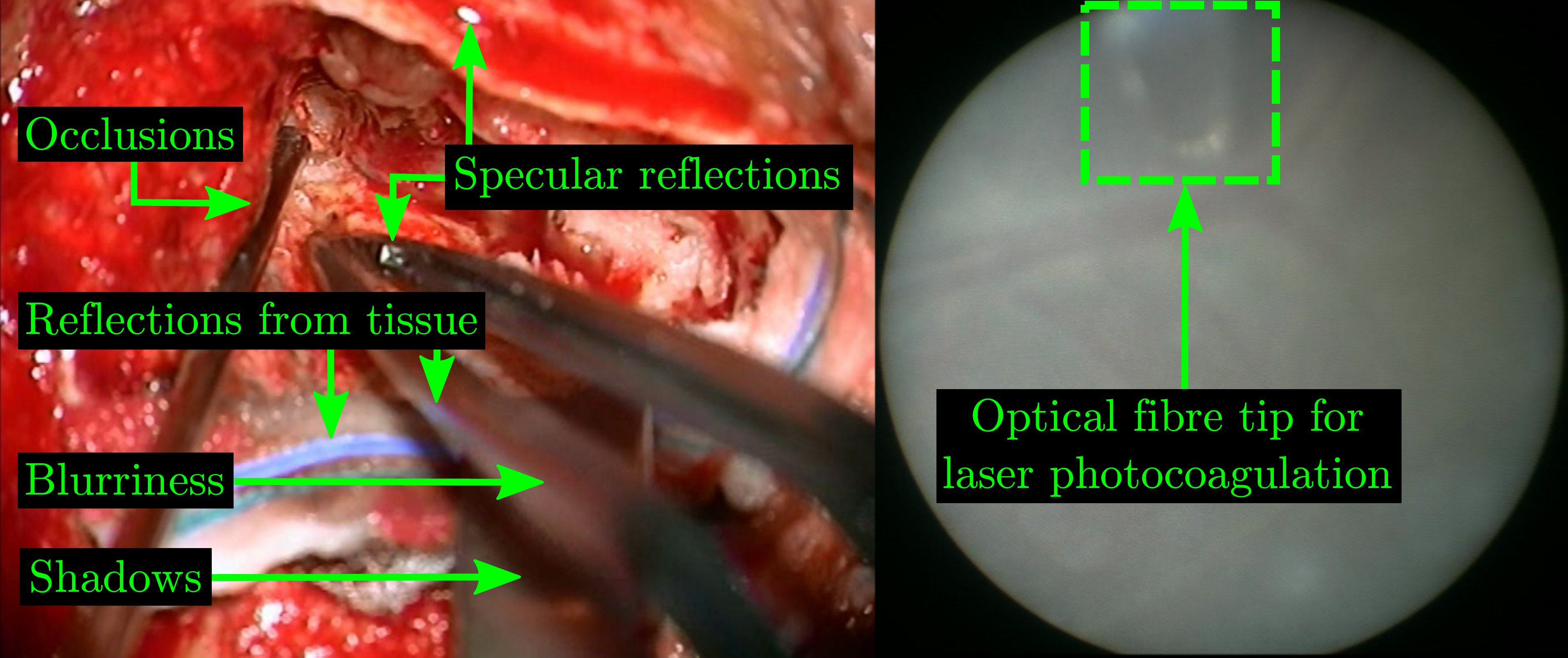}
	\caption{Challenges encountered by tool detection and localisation algorithms in real interventions. \textit{In vivo} neurosurgery~\cite{Bouget2015} (left). Twin-to-twin transfusion syndrome laser photocoagulation (right).}
	\label{fig:complications}
\end{figure}

There are a number of challenges that need to be addressed for real-time detection and tracking of surgical instruments. Endoscopic images typically present a vast amount of specular reflections (from both tissue and instruments), which is a source of confusion for segmentation algorithms as pixels that look the same belong to different objects (e.g. background and foreground). 
Changing lighting conditions, shadows and motion blur, combined with the complexity of the scene and the motion of organs in the background are also a challenge, as can be observed in \cref{fig:complications}.
As a result, anatomical structures and surgical instruments may look more similar than they actually are. Occlusions caused by body fluids and smoke also represent a major issue. Particularly for the case of fetal surgery, the turbidity of the amniotic fluid, makes the localisation of instruments really challenging, as can be observed in \cref{fig:complications}.
Fetal surgery also has the additional difficulty of relying on miniature endoscopes that contain several tens of thousands of fibres in an imaging guide. Transformed into pixels the number of fibres results in a very poor resolution (e.g. 30K in a \textsc{Karl Storz GmbH} 11508 AAK curved fetoscope~\cite{curvedFetoscope}).

To the best of our knowledge, in this paper, we present the first real-time ($\approx30$ fps) surgical tool segmentation pipeline. Our pipeline takes monocular video as input and produces a foreground/background segmentation based on both deep learning semantic labelling and optical flow tracking. The method is instrument-agnostic and can be used to segment different types of rigid or non-rigid instruments. We demonstrate that deep learning semantic labelling outperforms the state of the art on an open neurosurgical clinical dataset~\cite{Bouget2015}. Our results also show competitive performance between real-time and non-real-time implementations of our method.

\section{Methods}

\subsubsection{Convolutional-Neural-Network-based segmentation.}
There are several benefits of using a Convolutional Neural Network (CNN) compared to other state-of-the-art machine learning approaches~\cite{Bouget2015}. First, there is no need for trial and error to hand-craft features, as features are automatically extracted during the network training phase. As demonstrated in~\cite{Noh2015}, automatic feature selection does not negatively affect the segmentation quality. Furthermore, CNNs can be pre-trained on large general purpose datasets from the Computer Vision community and fine-tuned with a small amount of domain-specific images, as explained in~\cite{Long2014}. This particular feature of CNNs allows us to overcome the scarcity of labelled images faced by the CAI community. Therefore, it conveys the possibility of having an instrument segmentation mechanism that is not tool dependent, as demonstrated by our results.

\begin{figure}[b!]
	\centering
	\includegraphics[width=.8\textwidth]{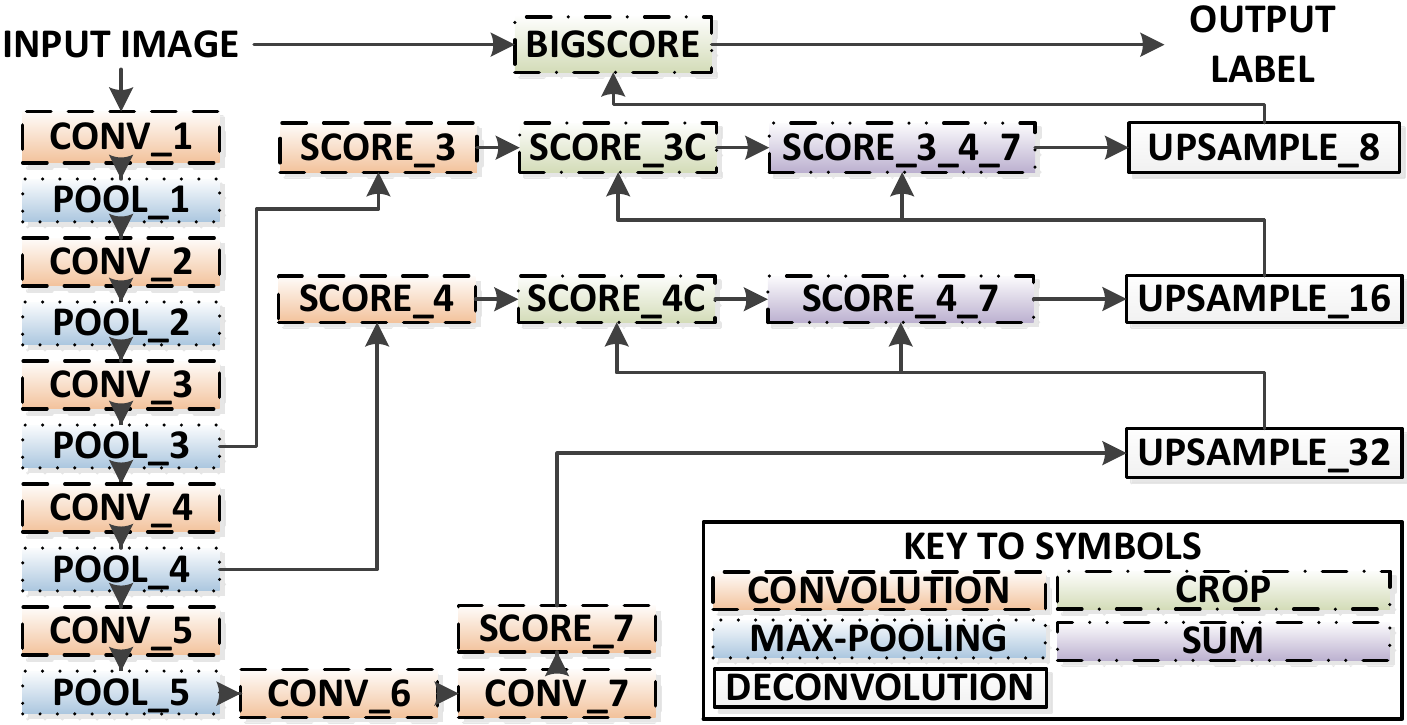}
	\caption{Illustration of the FCN-8s network architecture, as proposed in~\cite{Long2014}. In our method, the architecture of the network remains the same, but the number of outputs in SCORE\_3, SCORE\_4, SCORE\_5, UPSAMPLE\_8, UPSAMPLE\_16 and UPSAMPLE\_32 has been changed so that they produce only two scores per pixel, background and foreground.}
	\label{fig:cnn_diagram}
\end{figure}

Fully Convolutional Networks (FCN) are a particular type of CNN recently proposed by Long et al.~\cite{Long2014}. As opposed to previous CNNs such as AlexNet~\cite{alexnet2012} or VGG16~\cite{vgg16}, FCN are tailored to perform semantic labelling rather than classification. However, the two are closely related as FCN are built from adapting and fine-tuning pre-trained classification networks. In order to achieve this conversion from classification to segmentation two key steps are performed. First, the fully connected (FC) layers of the classification network are replaced with convolutions so that spatial information is preserved. Second, \textit{upsampling filters} (also called \textit{deconvolution layers}) are employed to generate a multi-class pixel-level output segmentation that features the same size of the input image. An essential characteristic of the \textit{upsampling filters} present in FCN is that their weights are not fixed, but initialised to perform bilinear interpolation and then learnt during the fine-tuning process. As a consequence, these networks are able to accept an arbitrary-sized input, produce a labelled output of equivalent dimensions and rely on end-to-end learning of labels and locations. That is, they behave as \textit{deep non-linear filters} that perform semantic labelling. There are three versions of the FCN introduced by Long et al., FCN-8s (shown in~\cref{fig:cnn_diagram}), FCN-16s and FCN-32s (available in the \textsc{Caffe} Model Zoo~\cite{modelzoo}). The difference between them being the use of intermediate outputs (such as the one coming from POOL\_3 or POOL\_4 in~\cref{fig:cnn_diagram}) in order to achieve finer segmentations.

\begin{figure}[b!]
	\begin{center}
		\includegraphics[width=\textwidth]{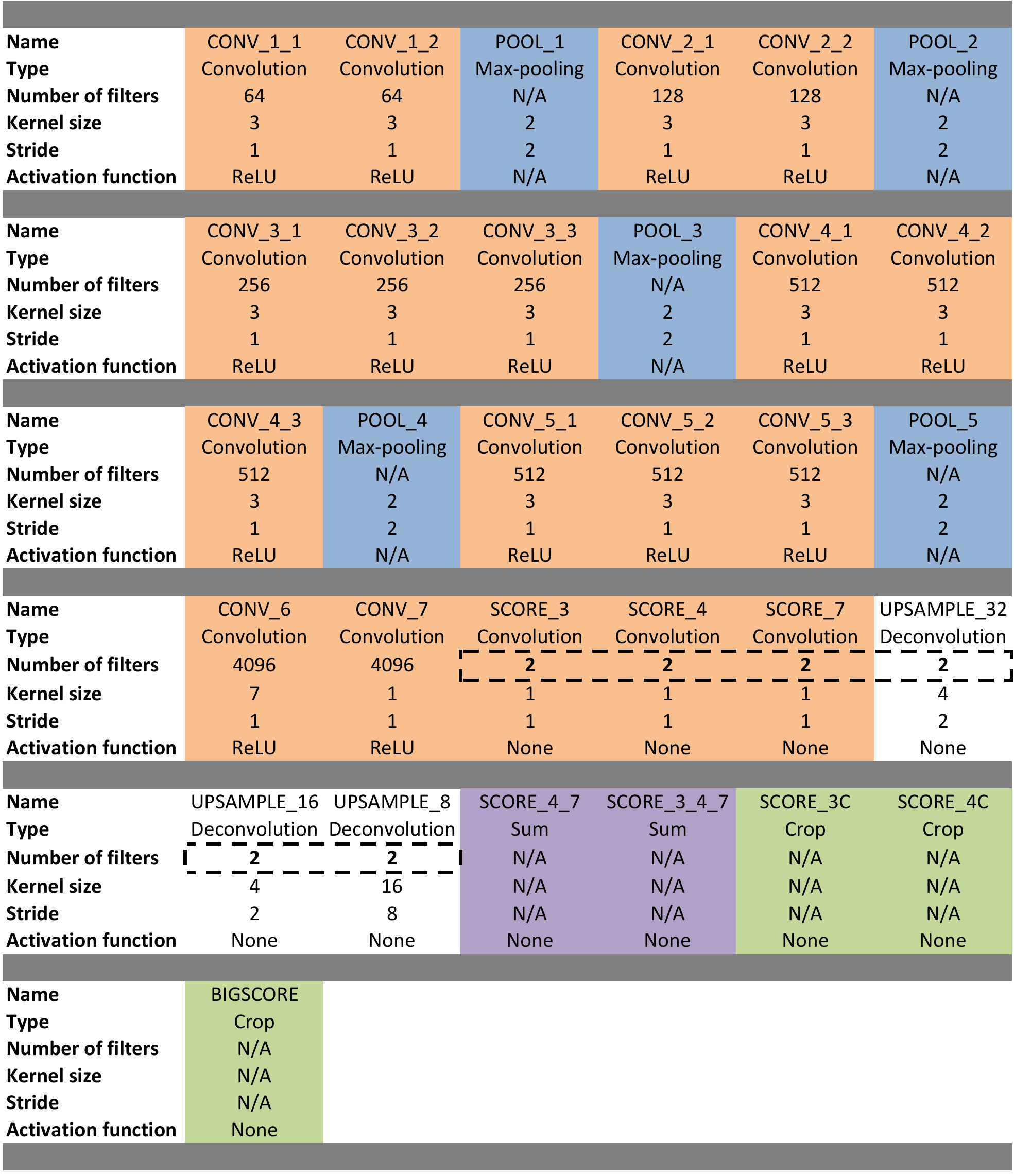}
		\caption{Parameters of the adapted FCN. Changes with respect to the original FCN-8s~\cite{Long2014} are shown surrounded by a dashed line.}
		\label{tab:cnn_table}
	\end{center}
\end{figure}

In this work, we have adapted and fine-tuned the FCN-8s~\cite{Long2014} for instrument segmentation. Its state-of-the-art performance in multi-class segmentation of general purpose computer vision datasets makes it a sensible choice for the task. The FCN-8s we employed was pre-trained on the PASCAL-context 59-class (60 including background)~\cite{mottaghi_cvpr14} dataset. As we are concerned with the separation of non-rigid surgical instruments from background, the structure of the network was adapted to provide only two scores per pixel by changing the number of outputs to just \textit{two} in the scoring and upsampling layers. This modification of parameters is highlighted within the dashed line in \cref{tab:cnn_table}. After this change, the network can be fine-tuned with a small amount of data belonging to a particular surgical domain. During inference, the final per-pixel scores provided by the FCN are normalised and calculated via \textit{argmax} to obtain per-pixel labels.

We have also implemented an improved learning process for the FCN. The optimiser selected to update the weights was the standard \textit{Stochastic Gradient Descent} (SGD). A key hyper-parameter of the fine-tuning process is the \textit{learning rate} (LR), which is the weight applied to the negative gradient used in the update rule of the optimisation. 
It has been recently shown in~\cite{Smith2015} that letting the learning rate fluctuate during the fine-tuning process achieves convergence to a higher accuracy in less number of iterations. This policy, introduced by Smith as Cyclical Learning Rate (CLR)~\cite{Smith2015}, may be implemented with different shapes (e.g. triangular, parabolic, sinusoidal). 
However, all of them produce similar results in~\cite{Smith2015}. We therefore choose the triangular window for the sake of simplicity.
As we are only interested in fine-tuning the network, the LR was constrained to a small value to tailor the parameters to the surgical domain without altering the behaviour of the network.
In our case, the LR boundaries, momentum and weight decay were set to [1e-13, 1e-10], 0.99 and 0.0005, respectively.

\subsubsection{Real-time segmentation pipeline.}
The drawback of the FCN we used is that it cannot run in real-time. \textsc{Caffe} performs forward evaluation in about 100ms for a 500$\times$500 RGB image using an \textsc{NVIDIA} GeForce GTX TITAN X GPU, but this computational time is well below the frame-rate of the endoscopic video, which is generally 25, 30, or 60 fps.

The key insight that was employed here to overcome this problem is that in the short time slot between two FCN segmentations, the tool remains roughly rigid and its appearance changes can be captured sufficiently well by an affine transformation. This type of transformation provides a trade-off between representing small changes and being robust enough for fast fitting purposes. Based on this assumption, tracking is used to detect the small motion between the last FCN-segmented frame and the current one. By registering the last FCN-segmented frame (as opposed to the most recently segmented frame) with the current one, we avoid the time-consuming feature point extraction in every frame and potentially reduce the propagation of error across frames.

\begin{figure}[b!]
	\includegraphics[width=\textwidth]{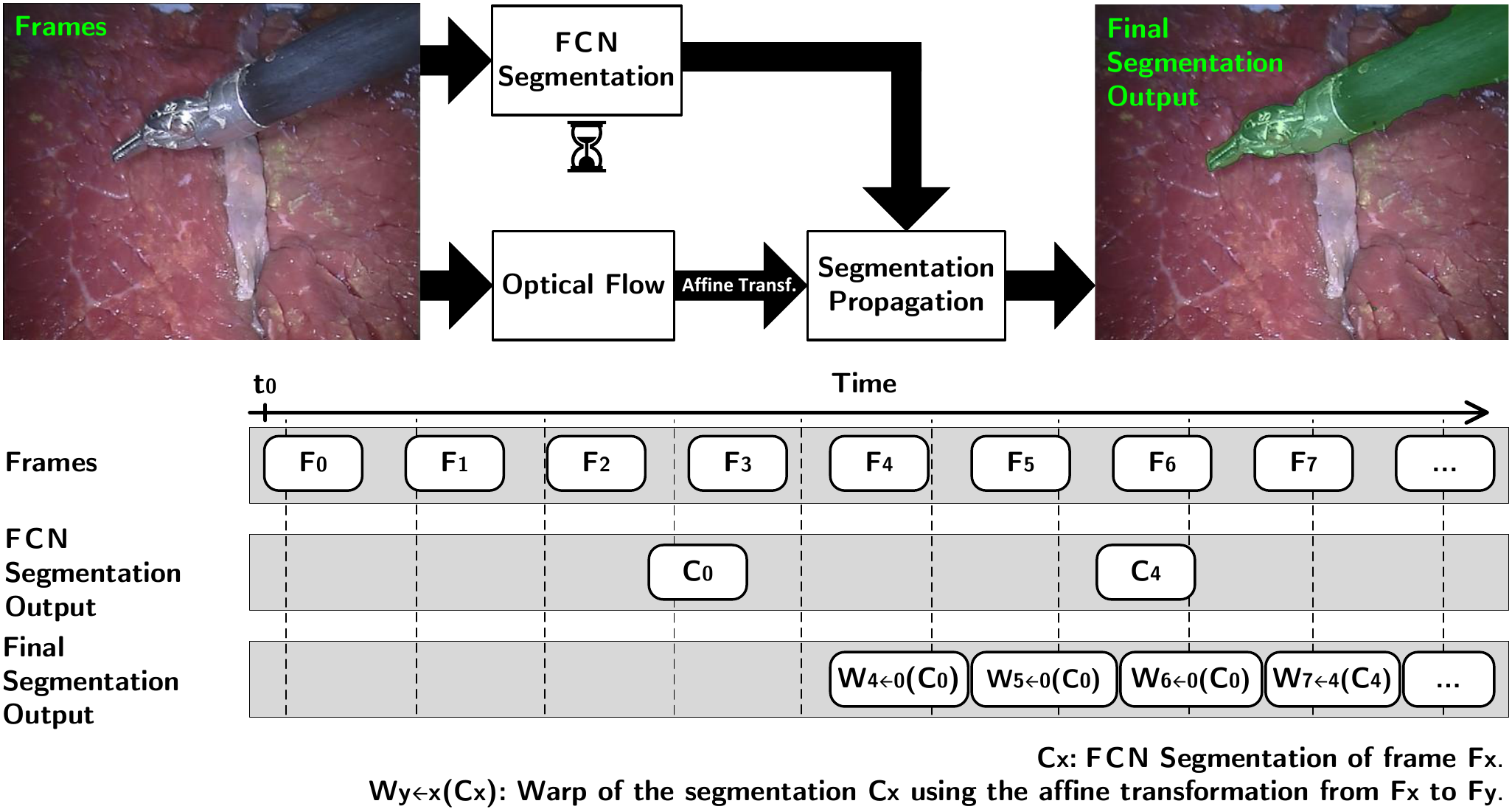}
	\caption{Real-time segmentation diagram and timeline. For the first few frames no FCN-based segmentation is available, hence the system does not provide any output. As soon as the first FCN output is retrieved, the system provides a segmentation per video frame. All the segmentation outputs \textbf{W} were obtained based on the last FCN-based output \textbf{C}.}
	\label{fig:timeline}
\end{figure}

Our asynchronous pipeline is illustrated in \cref{fig:timeline}. The FCN segmenter runs asynchronously to the rest of the pipeline. That is, when a frame is read from the video feed, it is sent to the FCN segmenter only if the FCN is not currently busy processing a previous frame. When the FCN finishes a segmentation, it updates the \textit{last} segmentation mask, which is stored in synchronised memory. Furthermore, the image just segmented is converted to grayscale (as matching feature points is faster than in colour images) and stored along with some (maximum 4000) foreground feature points for later use by the optical flow tracker. The feature points used are corners provided by the \texttt{GoodFeaturesToTrack} extractor (\textsc{OpenCV} implementation of the Shi-Tomasi corner detector~\cite{Tomasi1994}), which in combination with optical flow forms a widely successful tracking framework used for temporal constraints that satisfies our real-time requirement.
All the output segmentations are computed according to the following process. First, pyramidal Lukas-Kanade~\cite{Bouguet2002Pyramidal} optical flow is employed to find the correspondence between the foreground points in the previous FCN-segmented frame and the current received frame. Then the affine transformation between the two sets of points is estimated by solving the linear least squares problem
\begin{align*}
\textbf{\textit{A}}^{*},\,\t^{*} &:= \argmin_{\textbf{\textit{A}},\,\t} \Big( \sum_{i~\in~\textrm{inliers}}\norm{\n[i] - \textbf{\textit{A}}\,\p[i] - \t}^{2} \Big)
\end{align*}
with a RANSAC approach (\texttt{estimateRigidTransform}, \textsc{OpenCV} implementation to compute an optimal affine transformation between two 2D point sets) where $i$ is the iterator over the inlier feature-point matches, $\p$ is the set of points in the last FCN-segmented frame, $\n$ is the set of points in the frame that we are currently trying to segment and $[\textbf{\textit{A}}|\t]$ is the affine transformation between the two sets of points that we are estimating.

Once the affine transformation is obtained, it is applied to the \textit{last} segmentation mask produced by the FCN. This warped label is the final segmentation for the frame.
 
\section{Experiments and Results}

With the aim of demonstrating the flexibility of the presented methodology, three datasets have been used for validation. They contain training and test data for a wide variety of surgical settings, including \textit{in vivo} abdominal and neurological surgery and different set-ups of \textit{ex vivo} robotic surgery. Furthermore, they also contain different surgical instruments, i.e. rigid, articulated and flexible, respectively.  

\texttt{EndoVisSub}~\cite{MICCAIchallenge2015}. MICCAI 2015 Endoscopic Vision Challenge - Instrument Segmentation and Tracking Sub-challenge. This dataset consists of two sub-datasets, \textit{robotic} and \textit{non-robotic}. The training data for the \textit{robotic} sub-dataset is formed by four \textit{ex vivo} 45-second videos and the test data is formed by four 15-second and two 60-second videos. All of them having a resolution of 720$\times$576 and 25 fps. The training data for the \textit{non-robotic} sub-dataset is formed by 160 \textit{in vivo} abdominal images (coming from four different sequences) and the test data is formed by 4600 images (coming from nine different sequences). All of them having a resolution of 640$\times$480. No quantitative results are reported for the non-robotic \texttt{EndoVisSub} sub-dataset as ground-truth was not available from the challenge website.

\texttt{NeuroSurgicalTools}~\cite{Bouget2015}. This dataset consists of 2476 monocular images (1221 for training and 1255 for testing) coming from \textit{in vivo} neurosurgeries. The resolution of the images varies from 612$\times$460 to 1920$\times$1080.

\texttt{FetalFlexTool}. \textit{Ex vivo} fetal surgery dataset consisting of 21 images for training and a video sequence of 10 seconds for testing. In both the images and the video a non-rigid McKibben artificial muscle~\cite{Devreker2015} is actuated close to the surface of a human placenta. In order to prove the generalisation capabilities of the method, the training images were captured in air and the video was recorded under water, to facilitate different backgrounds and lighting conditions. The ground truth of both the training images and the testing video was produced through manual segmentation. The \textit{ex vivo} placenta used to generate this dataset was collected following a caesarean section delivery and after obtaining a written informed consent from the mother at University College London Hospitals (UCLH). The Joint UCL/UCLH Committees on Ethics of Human Research approved the study.

\begin{table}[b!]
	\centering
	\caption{Non-real-time quantitative results of the FCN-based segmentations. The results have been calculated based on the semantic labelling obtained for the testing images of each dataset. Three different FCN (one per dataset) have been fine-tuned to obtain these results.} 
	\begin{tabular}{lccc}
		\hline
		\multicolumn{1}{c}{\bfseries Dataset} & \multicolumn{1}{c}{\bfseries ~~Sensitivity~~} & \multicolumn{1}{c}{\bfseries ~~Specificity~~} & \multicolumn{1}{c}{\bfseries ~~Balanced Accuracy~~} \\ \hline
		\texttt{EndoVisSub (robotic)} & 72.2\% & 95.2\% & 83.7\% \\
		\texttt{NeuroSurgicalTools} & 82.0\% & 97.2\% & 89.6\% \\
		\texttt{FetalFlexTool} & 84.6\% & 99.9\% & 92.3\% \\ \hline
	\end{tabular}
	\vspace{0.2cm}
	\label{tab:cnnresults}
\end{table}

We implemented our method in C++, making use of the \textsc{Caffe-future} branch, acceleration from the NVIDIA CUDA Deep Neural Network library v4, using the Intel(R) Math Kernel Library as BLAS choice and the \textsc{CUDA} module of \textsc{OpenCV} 3.1. The results have been generated with an Intel(R) Xeon(R) (CPU) E5-1650 v3 @ 3.50GHz computer and a GeForce GTX TITAN X (GPU). All the results reported were obtained by fine-tuning the FCN for each dataset.

\begin{figure}[b!]
	\centering
	\includegraphics[width=\textwidth]{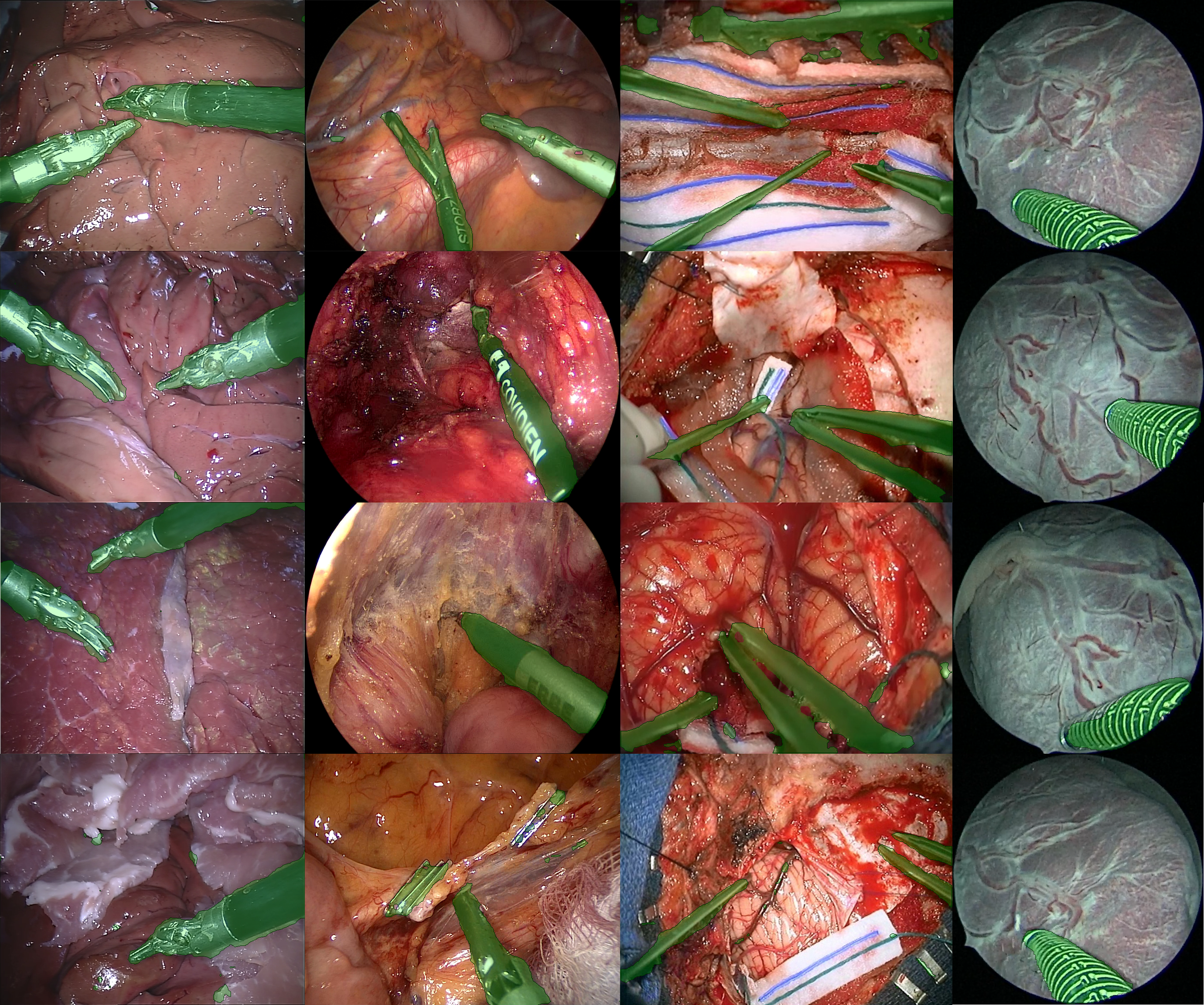}
	\caption{FCN-based segmentation of four testing images, each one belonging to a different dataset. From left to right, \texttt{EndoVisSub (robotic)}, \texttt{EndoVisSub (non-robotic)}, \texttt{NeuroSurgicalTools} (see~\cite{Bouget2015} Fig.5 for a qualitative comparison) and \texttt{FetalFlexTool}.}
	\label{fig:cnnqual}
\end{figure}

The first experiment carried out analysed the feasibility of FCN-based semantic labelling for instrument segmentation tasks without considerations for real-time requirements. The quantitative results can be seen in \cref{tab:cnnresults} and some segmentation examples are shown in \cref{fig:cnnqual} and the supplementary material. As can be seen in \cref{tab:cnnresults}, the balanced accuracy = (sensitivity + specificity) / 2 achieved for the \textit{in vivo} \texttt{NeuroSurgicalTools} dataset is 89.6\%, which is higher than the 85.8\% reported by~\cite{Bouget2015}.

\begin{figure}[b!]
	\centering
	\includegraphics[width=\textwidth]{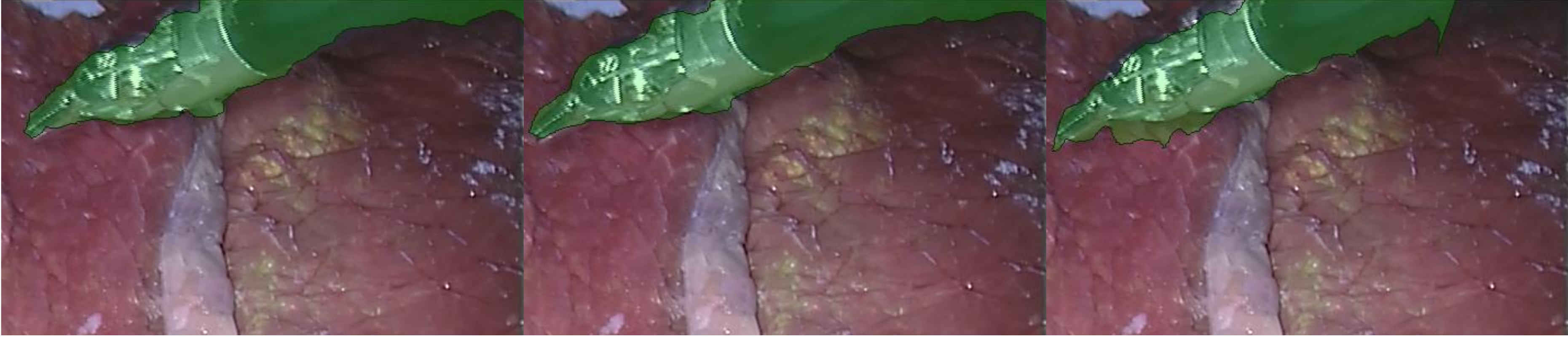}
	\caption{Comparison between FCN-based segmentation and tracking-based propagation. From left to right, previous frame segmented with FCN ($C_x$), current frame segmented with FCN ($C_y$) and tracking-based propagation ($W_{y\leftarrow x}(C_x)$).}
	\label{fig:c0c4w4}
\end{figure}

The real-time pipeline, including the mask propagation based on optical flow, was evaluated on \texttt{EndoVisSub (robotic)} and \texttt{FetalFlexTool} (no real-time results are reported for \texttt{NeuroSurgicalTools} due to lack of frame-by-frame video ground-truth). Quantitative results can be seen in~\cref{tab:realtime}. The real-time pipeline captures the tool with a performance which is acceptable in comparison to the off-line counterpart, as illustrated in~\cref{fig:c0c4w4} and the supplementary material. Our method was able to produce real-time ($\approx$ 30Hz) results for all the datasets.

\begin{table}[b!]
	\centering
	\caption{Quantitative results of the full real-time segmentation pipeline. The reported numbers are based on the frame-by-frame comparison of the binary labels provided by the presented real-time method and the ground truth video segmentations (for those datasets which have it).}
	\begin{tabular}{lccc}
		\hline
		\multicolumn{1}{c}{\bfseries Dataset} & \multicolumn{1}{c}{\bfseries ~~Sensitivity~~} & \multicolumn{1}{c}{\bfseries ~~Specificity~~} & \multicolumn{1}{c}{\bfseries ~~Balanced Accuracy~~} \\ \hline
		\texttt{EndoVisSub (robotic)} & 87.8\% & 88.7\% & 88.3\% \\
		\texttt{FetalFlexTool} & 36.3\% & 99.9\% & 68.1\% \\ \hline
	\end{tabular}
	\vspace{0.2cm}
	\label{tab:realtime}
\end{table}

\section{Discussion and Conclusion}

FCN stand out as a very promising technology for labelling endoscopic images. They can be fine-tuned with a small amount of medical images and no discriminative features have to be hand-crafted. Furthermore, these advantages are not at the expense of lowering the segmentation performance.

To the best of our knowledge this paper presents the first real-time FCN-based surgical tool labelling framework. Optical flow tracking can be successfully employed to propagate FCN segmentations in real-time. However, the quality of the results depends on how deformable the instruments being segmented are and how fast they move, as can be observed in the different results reported in \cref{tab:realtime}. The balanced accuracy achieved by the FCN-based labelling of the \texttt{EndoVisSub (robotic)} dataset (83.7\%) is lower than the one achieved by the real-time version (88.3\%). The increase in balanced accuracy from the FCN-based segmentation to the real-time version for the \texttt{EndoVisSub} is at the expense of a reduction in specificity. This is due to an inflation of the warped segmentation and related to the fact that several tools are present in the foreground and move in different directions. This may benefit the accuracy score by increasing sensitivity, similar effects have been observed for anchor box trackers (votchallenge.net). For the \texttt{FetalFlexTool} dataset which consists of a flexible McKibben actuator the balanced accuracy was reduced from 92.3\% to 68.1\%.

According to the results reported for the different datasets, we can conclude that the presented methodology is flexible enough to easily adapt to different clinical scenarios. Furthermore, feasibility for real-time segmentation of different surgical instruments has been demonstrated. This including non-rigid tools, as it is the case in the \texttt{FetalFlexTool} dataset.

However, as it would be expected, non-rigid foreground movements (either caused by the presence of several instruments or due to genuine non-rigid tool movements) that are faster than the time elapsed between two FCN segmentations (typically 100ms) affect the segmentation quality and will not be captured as well. This could be further addressed by separating the feature points detected on the foreground in different groups and using a set of affine transformations rather than a single one for the whole foreground.

Future work includes the possibility of detecting multiple instruments and also the inclusion of a Tracking Learning Detection framework~\cite{Kalal2010a}. At this stage, temporal information of previous segmentations is not fed to the FCN but is only used by the tracking system. It would be interesting to use long-term tracking information to both speed-up and improve the segmentation results.

\subsubsection{Acknowledgements.}
This work was supported by Wellcome Trust [WT101957], EPSRC (NS/A000027/1, EP/H046410/1, EP/J020990/1, EP/K005278), NIHR BRC UCLH/UCL High Impact Initiative and a UCL EPSRC CDT Scholarship Award (EP/L016478/1). The authors would like to thank NVIDIA for the donated GeForce GTX TITAN X GPU, their colleagues E. Maneas, S. Moriconi, F. Chadebecq, M. Ebner and S. Nousias for the ground truth of \texttt{FetalFlexTool} and E. Maneas for preparing setup with an \textit{ex vivo} placenta.

\bibliographystyle{splncs}
\bibliography{library}

\end{document}